\documentclass[11pt]{article}

\usepackage[final]{acl}

\usepackage{times}
\usepackage{latexsym}
\usepackage{booktabs}
\usepackage{multirow}
\usepackage{tcolorbox}
\usepackage[T1]{fontenc}

\usepackage[utf8]{inputenc}

\usepackage{microtype}

\usepackage{inconsolata}

\usepackage{graphicx}
\usepackage{subcaption}

\usepackage{amssymb}
\usepackage{amsmath}
\usepackage{xcolor}

\title{\emph{IntroLM}: Introspective Language Models via Prefilling-Time Self-Evaluation}

\author{
Hossein Hosseini Kasnavieh$^{1,2}$ \quad
Gholamreza Haffari$^{3}$ \quad
Chris Leckie$^{2}$ \quad
Adel N. Toosi$^{1,2}$ \\
$^{1}$DisNet Lab, School of Computing and Information Systems, The University of Melbourne, Australia \\
$^{2}$School of Computing and Information Systems, The University of Melbourne, Australia \\
$^{3}$Department of Data Science \& AI, Monash University, Australia \\
\texttt{hossein.hosseini@student.unimelb.edu.au} \\
\texttt{\{caleckie, adel.toosi\}@unimelb.edu.au} \\
\texttt{gholamreza.haffari@monash.edu}
}

\begin{document}
\maketitle
\begin{abstract}
A major challenge for the operation of large language models (LLMs) is how to predict whether a specific LLM will produce sufficiently high-quality output for a given query.  Existing approaches rely on external classifiers, most commonly BERT-based models, which suffer from limited context windows, constrained representational capacity, and additional computational overhead. We propose IntroLM, a method that enables causal language models to predict their own output quality during the prefilling phase without affecting generation using \texttt{[CPX]} tokens. By introducing token-conditional LoRA that activates only for the introspective \texttt{[CPX]} token, the model learns to predict the output quality for a given query while preserving the original backbone behavior and avoiding external evaluators. On question-answering benchmarks, IntroLM applied to \textsc{Qwen3-8B} achieves a ROC–AUC of 90\% for success prediction, outperforming a \textsc{DeBERTa-v3-Large} classifier by 14\%. When integrated into multi-model routing systems, IntroLM achieves superior cost–performance trade-offs, reducing end-to-end latency by up to 33\% and large-model usage by up to 50\% at matched reliability. Our code is available at \url{https://github.com/hhosseini1377/LLM_routing}.
\end{abstract}
\section{Introduction}

Large Language Models (LLMs) have demonstrated exceptional capabilities across tasks ranging from complex reasoning to code generation. The model landscape has grown remarkably diverse, spanning proprietary systems like GPT \citep{OpenAI2025} and Claude \citep{anthropic2024} alongside open-source alternatives such as LLaMA \citep{Llama4_2025}, Qwen \citep{yang2025qwen3}, and Mistral \citep{Mistral3_2025}. Each of these model families offers variants in various sizes—ranging from a few billion to hundreds of billions of parameters—with distinct tradeoffs between computational cost and performance. Broadly speaking, smaller models tend to perform well on simpler prompts, while larger, more capable models excel on complex tasks requiring sophisticated reasoning at higher computational costs.

Predicting a model’s output quality on a given prompt prior to generation is crucial for effective LLM deployment. Such estimates enable confidence scoring, intelligent prompt routing in multi-model systems, and selective computation to optimize resource usage \citep{ongroutellm, chenfrugalgpt}. In routing-based deployments, output quality prediction allows simpler prompts to be handled by smaller, more efficient models, while more complex prompts are escalated to larger models with stronger reasoning capabilities, thereby balancing accuracy and cost without incurring unnecessary computational overhead.

Prior routing studies predominantly rely on BERT-based classifiers \cite{devlin2019bert} and are typically evaluated on chat-style datasets with short prompts \citep{zooter, chenfrugalgpt, best-route, hybridllm, ongroutellm}. For example, UltraChat \citep{ultrachat} has average prompt lengths below 200 tokens. In contrast, modern LLM deployments prepend substantial reference material, yielding prefilling inputs that can span tens or hundreds of thousands of tokens, well beyond BERT’s fixed 512-token context window, thereby limiting the applicability of BERT-based routers to long-context complexity evaluation.

These limitations motivate the use of more capable models for prompt complexity evaluation, but relying on a separate large model introduces substantial computational overhead. In both confidence estimation and routing settings, this approach requires additional model inference solely for evaluation, which can negate the efficiency gains that complexity-aware systems are designed to achieve. This creates a fundamental tension: accurate complexity prediction benefits from stronger models, yet deploying separate evaluators undermines overall system efficiency.

\begin{figure}[t]
  \includegraphics[width=\columnwidth]{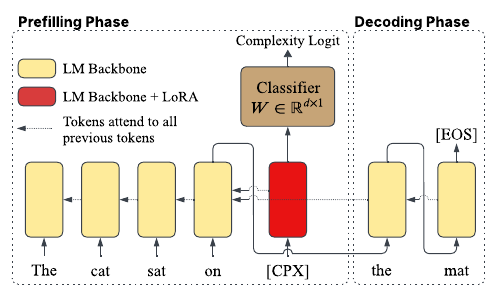}
  \caption{Overview of IntroLM. Complexity logits are computed during prefilling using \texttt{[CPX]} tokens, while generation remains unchanged.}
  \label{fig:IntroLM_base}
\end{figure}

To address these challenges, we introduce \textit{Introspective Language Models (IntroLM)}, a method that enables causal language models to utilize their backbone representations to predict prompt difficulty during the prefilling phase, without relying on external evaluators or altering generation behavior. Figure~\ref{fig:IntroLM_base} provides an overview of the IntroLM architecture. IntroLM appends special \texttt{[CPX]} introspection tokens to the input, whose hidden states are used to estimate the probability that the model will successfully answer the prompt. Crucially, \texttt{[CPX]} tokens are excluded from the key–value cache and are not used to initialize decoding, ensuring that autoregressive generation begins from the original prompt and that decoding dynamics and output distributions remain identical to those of the base model. To adapt the model for this introspective task while preserving generative behavior, we introduce token-conditional LoRA, which selectively modifies representations associated with \texttt{[CPX]} tokens during prefilling.

IntroLM offers several practical deployment advantages. It eliminates the need for a separate evaluator model, requiring only lightweight LoRA parameters, and reuses backbone representations computed during prefilling, avoiding redundant computation on long-context inputs. Moreover, complexity estimation incurs no additional decoding or prompt engineering and naturally operates over the full prefilling input, enabling low-latency decision-making. In our evaluations, we emphasize long-context question-answering benchmarks, including HotpotQA \citep{hotpotqa}, to assess prefilling-time complexity under reference-heavy, multi-hop reasoning inputs. Across these benchmarks, IntroLM applied to \textsc{Qwen3-8B} achieves a ROC--AUC of 90\%, outperforming a \textsc{DeBERTa-v3-Large} classifier by 14 points.

As a practical application, we incorporate IntroLM into multi-model routing. We compare IntroLM-based routing against a standard BERT-based external router and show that introspective, prefilling-time signals enable more efficient routing decisions. Across two routing scenarios, IntroLM-based routing reduces the fraction of prompts escalated to the larger model by up to 50\% and lowers end-to-end latency by up to 33\% at matched reliability.



\section{Related Work}
Prior work on output quality estimation and routing can be broadly categorized by when quality is assessed. Some approaches evaluate quality after execution by generating responses and verifying them post hoc, incurring additional computation and latency \citep{chenfrugalgpt, automix}. Other methods perform pre-execution routing, predicting output quality before running the prompt, typically using external evaluators such as BERT-style encoder classifiers \citep{hybridllm, best-route, ongroutellm, zooter}. While these approaches avoid executing multiple models at inference time, they rely on encoders with fixed and limited context windows (e.g., 512 tokens), making them ill-suited for modern long-context settings such as retrieval-augmented generation. Closest to our work, \citet{confidence} introduce special confidence tokens that are generated at the end of decoding to infer model confidence; in contrast, IntroLM performs introspection during the prefilling phase, enabling quality estimation before any generation. A detailed discussion of related studies is provided in Appendix~\ref{app:related}.

\section{Preliminaries}

\subsection{Problem Formulation}
Let $M : \mathcal{X} \rightarrow \mathcal{Y}$ be a language model that maps a
prompt $x \in \mathcal{X}$ to an output $\hat{y} = M(x)$. Our main goal is to estimate
the probability that the model produces a high-quality output for a given
prompt, formalized as $P(\ell = 1 \mid x)$, where $\ell \in \{0,1\}$ is a binary
label indicating whether the output satisfies a predefined task-dependent
quality criterion (e.g., correctness). Throughout this work, we assume $M$ is
fixed and omit it from the conditioning for brevity.

Given a dataset $\mathcal{D} = \{(x_i, \ell_i)\}_{i=1}^N$, where each prompt $x_i$
is labeled according to whether the corresponding model output
$\hat{y}_i = M(x_i)$ is high quality ($\ell_i = 1$) or not ($\ell_i = 0$), we seek to
learn a complexity evaluator:
\[
f_\theta : \mathcal{X} \rightarrow [0,1],
\]
which estimates the probability $P(\ell = 1 \mid x)$ based solely on the prompt,
without requiring access to the generated output. While we focus on binary
supervision in this work, the formulation naturally extends to ordinal or
continuous quality scores (e.g., multi-level ratings) by replacing $\ell_i$ with
graded supervision. We train $f_\theta$ by minimizing the cross-entropy loss.

\subsection{Prefilling and Decoding Stages of LLMs}
Causal language models process input through two distinct phases:
\textbf{prefilling} and \textbf{decoding}. During the prefilling phase, the
model processes the entire prompt in parallel, computing hidden states for each
token through self-attention. In this mechanism, each token attends to all
preceding tokens in the sequence, allowing information to aggregate across the
entire prompt prefix. As a result, the hidden state of the final token naturally
encodes a contextualized representation of the full input sequence. The
decoding phase then commences from this representation and generates
subsequent tokens autoregressively \citep{naveed2025comprehensive}.

Motivated by the unidirectional attention structure of decoder-only models, one might attempt to perform complexity evaluation during prefilling by applying a classifier to the final hidden state of the prompt. While simple, this approach yields suboptimal performance in practice, as backbone representations are optimized for autoregressive generation rather than complexity assessment. Moreover, directly fine-tuning the backbone to improve prediction would alter generation behavior, which is undesirable. We empirically confirm these limitations in Section~\ref{cpx_impact}, especially for more complex tasks. In the next section, we propose our technique to adapt the prompt complexity task without affecting the generative phase.

\begin{figure}[t]
  \includegraphics[width=\columnwidth]{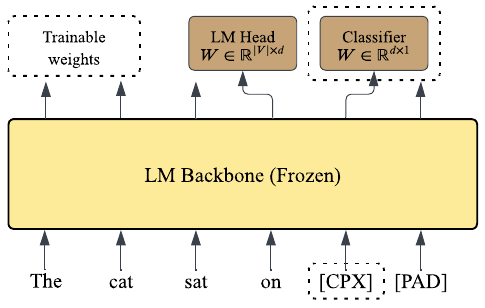}
  \caption{Training via Frozen Weights for [CPX]. Only the classifier head and \texttt{[CPX]} token embedding are trainable.}
  \label{fig:training_base}
\end{figure}

\section{Introspective Language Models (IntroLM)}
To address the mismatch between representations optimized for generation and those required for complexity evaluation, we introduce special \texttt{[CPX]} tokens appended to the end of the Because these tokens appear after the full
prompt, each \texttt{[CPX]} token attends to the entire prefix during the
prefilling phase, enabling the model to construct representations dedicated to
complexity estimation. A lightweight classifier head is applied to the hidden
states of these tokens to produce a complexity score.

Crucially, representations used for generation remain unchanged: autoregressive decoding is initialized from the last hidden state of the final token in the original prompt (excluding \texttt{[CPX]}), and backbone parameters governing the prompt tokens are kept frozen.
In addition, the \texttt{[CPX]} token is excluded from the key--value cache used
during decoding, ensuring that generated tokens cannot attend to
\texttt{[CPX]} representations. This separation allows the model to learn
task-specific complexity signals without altering its generation behavior. The overall structure of our method is illustrated
in Figure~\ref{fig:IntroLM_base}. In the following, we propose two methods for training IntroLM.


\subsection{Training via Frozen Weights for \texttt{[CPX]}}

As a first training approach, we consider a minimal-intervention approach in
which the entire model backbone is frozen and only the classifier head and the
\texttt{[CPX]} token embedding are trained. In this setting, the
\texttt{[CPX]} token acts as a learned soft prompt: during prefilling, it
attends to the original prompt tokens through the frozen attention mechanism
and aggregates a contextualized hidden-state representation. This representation
is then passed to a lightweight linear classifier to predict the complexity
score.

This design preserves the original generative behavior of the model by
restricting all parameter updates to the introspection pathway. At the same
time, it highlights the extent to which complexity estimation can be achieved
using representations induced by a fixed backbone. In the following, we build
on this formulation by introducing a more expressive adaptation mechanism that
selectively updates the computation paths seen by \texttt{[CPX]} tokens, while
maintaining the same separation between introspection and generation.
Figure~\ref{fig:training_base} illustrates this baseline training setup.

\begin{figure}[t]
\centering
  \includegraphics[width=\columnwidth]{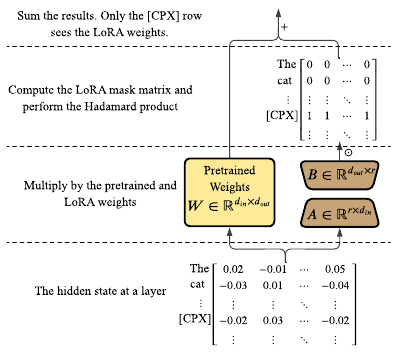}
  \caption{An illustration of token-conditional LoRA in any projection. The updated LoRA weights are effectively masked for non-\texttt{[CPX]} tokens.}
  \label{fig:token_conditioned}
\end{figure}

\subsection{Training via Token-Conditional LoRA}
Low-Rank Adaptation (LoRA) \citep{hulora} enables efficient fine-tuning of large language models by injecting trainable low-rank matrices into the attention and feed-forward layers. By decomposing weight updates as $\Delta W = BA$ where $B \in \mathbb{R}^{d \times r}$ and $A \in \mathbb{R}^{r \times d}$ with rank $r \ll d$, LoRA achieves parameter-efficient adaptation with minimal computational overhead compared to full fine-tuning.

We introduce \emph{token-conditional LoRA}, a parameter-efficient adaptation
mechanism that applies low-rank updates selectively to specific tokens while
leaving all others unaffected.
Figure~\ref{fig:token_conditioned} provides a schematic overview of this
mechanism. For clarity, the figure illustrates the operation for a single input
sequence, while the formulation below is presented for batched inputs.

Concretely, consider a generic linear projection
\( W \in \mathbb{R}^{d_{\text{in}} \times d_{\text{out}}} \) applied to hidden
states \( H \in \mathbb{R}^{B \times n \times d_{\text{in}}} \).
We compute both the standard transformation \( HW \) and a LoRA update
\( H\Delta W \), where \( \Delta W = BA \).

To restrict adaptation to introspection tokens, we construct a binary mask
\( M \in \mathbb{R}^{B \times n \times d_{\text{out}}} \) whose entries are $1$
at \texttt{[CPX]} token positions and $0$ elsewhere.
The final output is computed as
\[
HW + (H\Delta W) \odot M,
\]
where \( \odot \) denotes element-wise multiplication.
This ensures that only \texttt{[CPX]} tokens receive LoRA-modified
representations, while all other tokens follow the pretrained backbone. This formulation applies uniformly to all transformer projections, including
both attention and feed-forward layers, preserving generation behavior while
enabling localized adaptation for complexity evaluation. 

\textit{IntroLM with token-conditional LoRA} introduces negligible runtime and
parameter overhead. Complexity evaluation is performed during the prefilling
phase, where the added \texttt{[CPX]} token are processed in parallel with the
original prompt tokens, and decoding proceeds unchanged. The classifier head is
lightweight, and the additional LoRA parameters account for less than 1\% of the
model size.

\paragraph{Impact on projections and \texttt{[CPX]} adaptation.}
Modern LLMs employ transformer blocks composed of multi-head self-attention and
gated feed-forward networks. The attention mechanism transforms hidden states
$H \in \mathbb{R}^{n \times d}$ through query ($Q$), key ($K$), and value ($V$)
projections, computing $\text{softmax}(QK^{\top}/\sqrt{d})$ to produce
contextualized representations. The gated feed-forward network applies parallel
gate and up projections, followed by element-wise gating and a down projection:
\[
\text{FFN}(x) = W_{\text{down}} \cdot \big(\sigma(W_{\text{gate}} \cdot x)
\odot (W_{\text{up}} \cdot x)\big),
\]
where $\sigma$ denotes a non-linear activation (typically SiLU) and $\odot$
indicates element-wise multiplication. In total, each transformer block
typically contains seven distinct linear projections \citep{yang2025qwen3}.

Token-conditional LoRA enables selective adaptation of the computation paths
seen by \texttt{[CPX]} tokens during prefilling. In particular, adapting the
query projection allows \texttt{[CPX]} to learn how to attend to different parts
of the original prompt. The key and value projections, which are produced by the
original prompt tokens and play a dominant role in determining the attention
pattern, are kept frozen and are therefore omitted from the LoRA targets.
Empirically, adapting only the query projection (for the attention block) is sufficient to enable effective
introspective behavior (see Section~\ref{LoRA_effects}). In addition, the feed-forward network projections
(\texttt{gate}, \texttt{up}, and \texttt{down}) seen by \texttt{[CPX]} tokens can
be adapted to further refine complexity-specific representations, while
preserving the behavior of the original prompt tokens.

\section{IntroLM-Based Routing}

\subsection{Routing  Problem}
We consider a routing setting with access to two language models: a smaller,
lower-cost model $M_s$ and a larger, higher-capacity model $M_\ell$. In typical
deployments, $M_s$ may correspond to a locally hosted open-source model, while
$M_\ell$ represents a proprietary model that incurs substantially higher
computational or monetary cost.

Given a prompt $x \in \mathcal{X}$, a routing strategy selects an execution
pathway—potentially involving partial computation—such that the final generated
output attains a desired level of performance. The objective is to minimize
unnecessary use of $M_\ell$ while maintaining reliable task performance. In this
work, routing decisions are based on a scalar \emph{capability score} in $[0,1]$,
where larger values indicate higher confidence that the small model $M_s$ can
successfully handle the prompt.

\subsection{Our Routing Approach}
We propose a \emph{prefill-aware} routing strategy that leverages IntroLM to
estimate the capability of the small model using internal signals obtained
during the prefilling phase. Given a prompt $x \in \mathcal{X}$, the system first
executes prefilling on $M_s$, producing cached key--value states required for
decoding, along with a capability score
$f_{\text{IntroLM}}(x) \in [0,1]$ that estimates the probability that $M_s$ can
produce a correct output for $x$.

The routing policy $\pi^{\alpha}_{\text{IntroLM}} : \mathcal{X} \rightarrow
\{M_s, M_\ell\}$ is defined as
\[
\pi^{\alpha}_{\text{IntroLM}}(x) =
\begin{cases}
M_s, & \text{if } f_{\text{IntroLM}}(x) \ge \alpha, \\
M_\ell, & \text{otherwise}.
\end{cases}
\]\label{threshold}
If $\pi^{\alpha}_{\text{IntroLM}}(x)=M_s$, generation proceeds by decoding on
$M_s$ using the cached prefilling states without recomputation. If the prompt is
escalated, it is reprocessed by $M_\ell$, which performs both prefilling and
decoding. Increasing the threshold $\alpha$ yields more conservative routing by
escalating a larger fraction of prompts to $M_\ell$, trading off computational
cost for higher reliability.

\subsection{Routing Metrics}
\label{sec:metrics}
Routing strategies induce different trade-offs between output reliability and computational cost.
To quantify these trade-offs, we evaluate routing policies as a function of the threshold $\alpha$ using
one performance metric, \emph{reliability}, and two cost metrics, \emph{large-model call rate} and
\emph{end-to-end latency}.

\begin{figure*}[t]
    \centering
    \begin{subfigure}[t]{0.66\columnwidth}
        \centering
        \includegraphics[width=\linewidth]{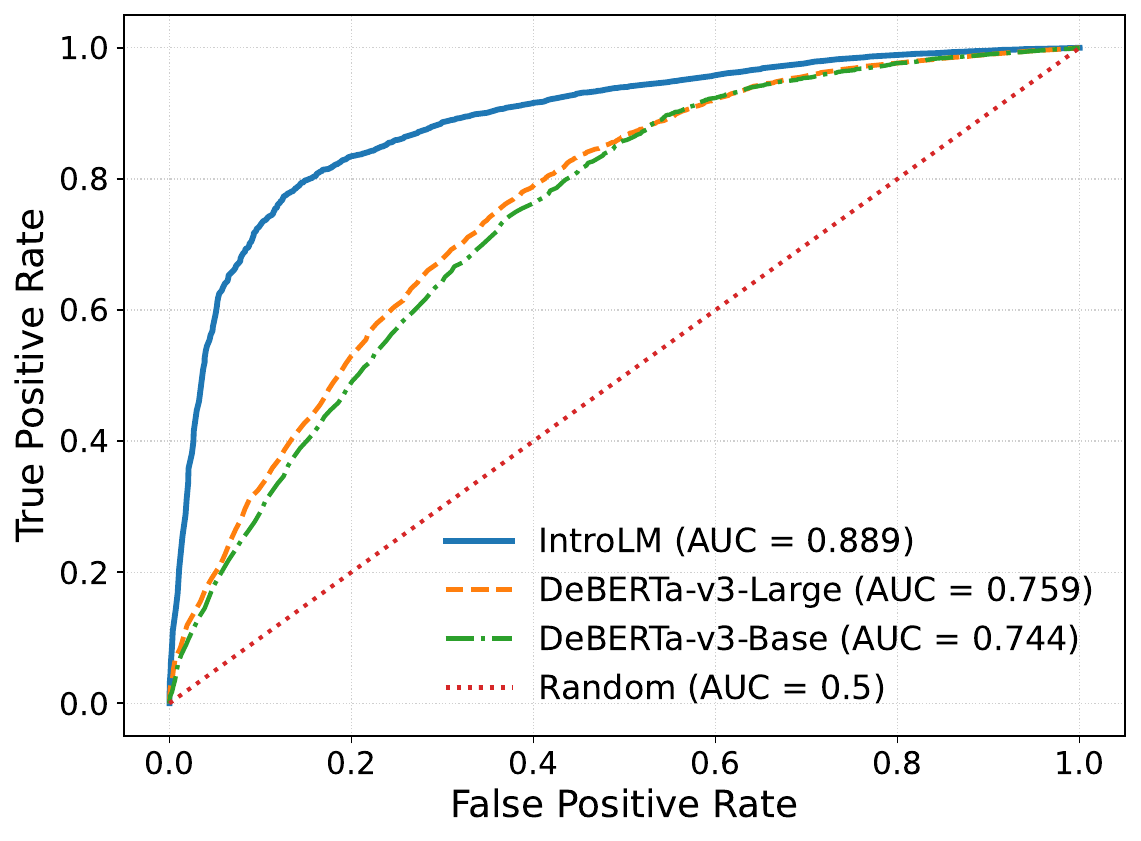}
        \caption{\textsc{General QA}}
        \label{fig:roc_general_qa}
    \end{subfigure}
    \hfill
    \begin{subfigure}[t]{0.66\columnwidth}
        \centering
        \includegraphics[width=\linewidth]{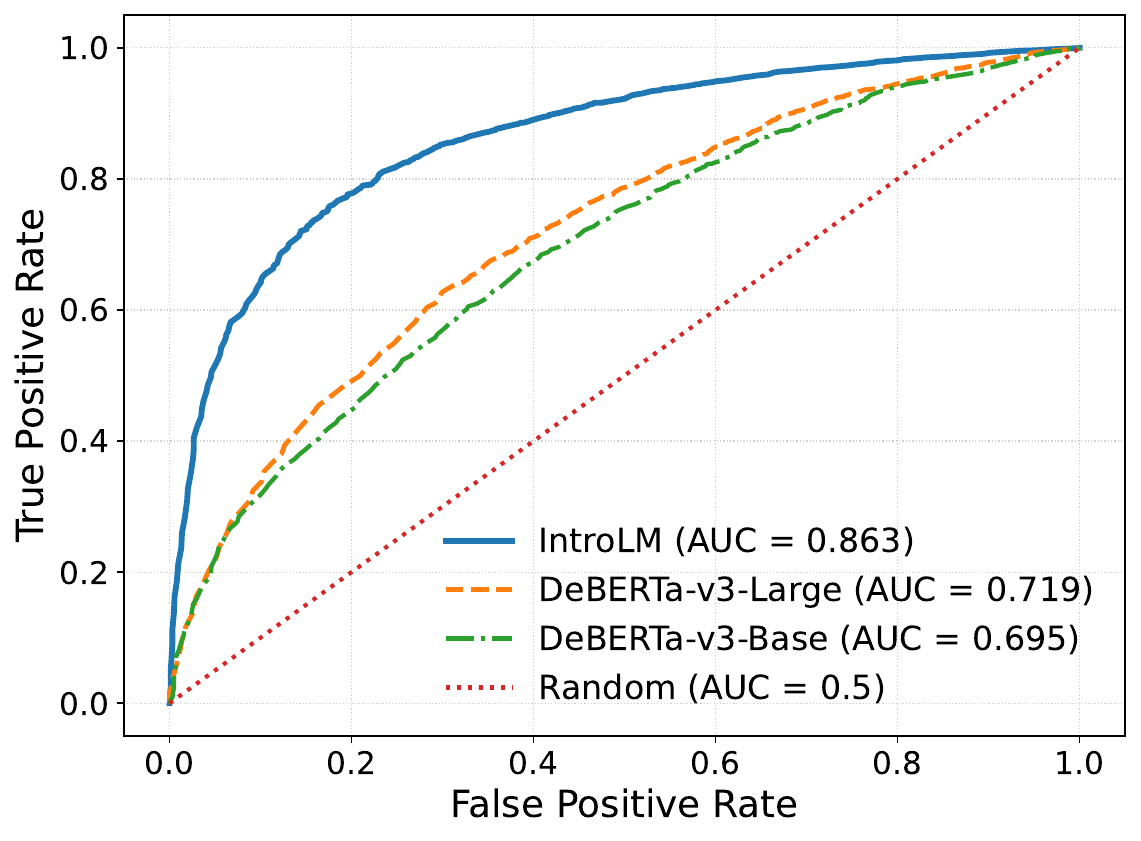}
        \caption{\textsc{HotpotQA}}
        \label{fig:roc_hotpotqa}
    \end{subfigure}
    \hfill
    \begin{subfigure}[t]{0.66\columnwidth}
        \centering
        \includegraphics[width=\linewidth]{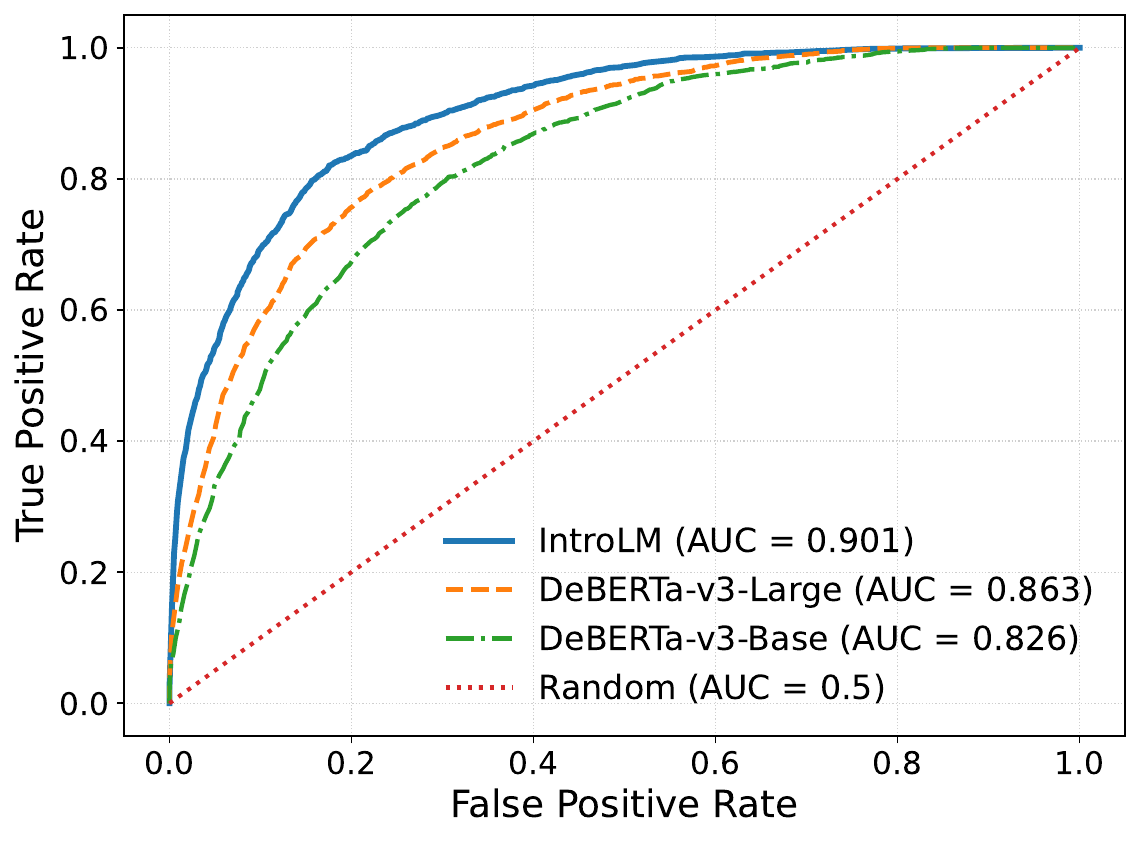}
        \caption{\textsc{Daily-style chat}}
        \label{fig:roc_hotpotqa}
    \end{subfigure}
    \caption{ROC curves comparing IntroLM (solid blue curve) and BERT-based baselines on
(a) General QA (b) HotpotQA (c) daily-style chat prompts (LMSYS-
Chat-1M). IntroLM achieves consistently stronger
separability between simple and complex prompts.
}
    \label{fig:roc_curves}
\end{figure*}

\paragraph{Reliability}
We define the reliability of a routing policy $\pi^{\alpha}$ as the probability that the final output
satisfies the quality criterion. Formally, letting $\ell~\in~\{0,1\}$ denote the ground-truth success
label for prompt $x$, reliability is defined as
\[
R(\pi^{\alpha})
=
1 -
\mathbb{E}_{(x,\ell)\sim\mathcal{D}}
\big[
\mathbf{1}\{\pi^{\alpha}(x)=M_s\}\,(1-\ell)
\big].
\]
Throughout our routing evaluation, we treat the large model $M_\ell$ as a
high-reliability fallback. This simplifying assumption allows us to isolate
the behavior of the routing mechanism by attributing observed failures to
incorrect executions on the smaller model $M_s$. Under this formulation,
reliability reflects the router’s ability to avoid routing prompts that $M_s$
cannot successfully handle, which is the dominant source of quality
degradation in cost-aware routing systems.

\paragraph{Large-model call rate}
The large-model call rate measures the expected fraction of prompts routed to $M_\ell$:
\[
c_\ell(\pi^{\alpha})
=
\mathbb{E}_{x\sim\mathcal{D}}
\big[
\mathbf{1}\{\pi^{\alpha}(x)=M_\ell\}
\big].
\]
This metric is particularly informative in settings where the cost of invoking the large model is
substantially higher than that of the smaller model, such as when routing between locally hosted
open-source models and proprietary, high-cost APIs.

\paragraph{Latency}
For a model $M$, time-to-first-token (TTFT) is defined as the time required to complete prefilling
and emit the first output token, and the time-per-output-token (TPOT) as the average time to generate
each subsequent token \citep{stojkovic2025dynamollm}. For an average output length $L$, the expected
end-to-end latency is:
\[
T_M(L) = \mathrm{TTFT}_M + (L-1)\,\mathrm{TPOT}_M .
\]

\noindent\emph{BERT-based routing.}
The expected latency is:
\begin{align*}
T^{\alpha}_{\text{BERT}}(L)
&=
\big(1 - c_\ell(\pi^{\alpha}_{\text{BERT}})\big)\,T_{M_s}(L)
\\
&\quad
+\, c_\ell(\pi^{\alpha}_{\text{BERT}})\,T_{M_\ell}(L).
\end{align*}

\noindent\emph{IntroLM-based routing.}
Since prefilling is always executed on $M_s$, the expected latency is
\begin{align*}
T^{\alpha}_{\text{IntroLM}}&(L)
=
\mathrm{TTFT}_{M_s}
\\
&\quad
+ \big(1 - c_\ell(\pi^{\alpha}_{\text{IntroLM}})\big)\,(L-1)\,\mathrm{TPOT}_{M_s}
\\
&\quad
+\, c_\ell(\pi^{\alpha}_{\text{IntroLM}})\,T_{M_\ell}(L).
\end{align*}

This latency-based evaluation is particularly relevant when both models are
hosted locally, where end-to-end response time is the dominant cost factor.
\section{Experiments}

\subsection{Experimental Setup}

\paragraph{Datasets}
We evaluate complexity prediction under three complementary settings:
(i) \textsc{General QA}, combining \textsc{MMLU} \citep{mmlu},
\textsc{MMLU-Pro} \citep{mmlu-pro}, and \textsc{GSM8K} \citep{gsm8k};
(ii) \textsc{HotpotQA} \citep{hotpotqa}, a long-context, multi-hop reasoning
benchmark; and (iii) \textsc{LMSYS-Chat-1M}, a large-scale collection of real-world
chat prompts used to evaluate daily-style conversational inputs. Full details on dataset
construction, filtering, labeling, and prompt templates are provided in
Appendix~\ref{app:datasets}.

\begin{figure*}[t]
    \centering

    \begin{subfigure}[t]{0.24\textwidth}
        \centering
        \includegraphics[width=\linewidth]{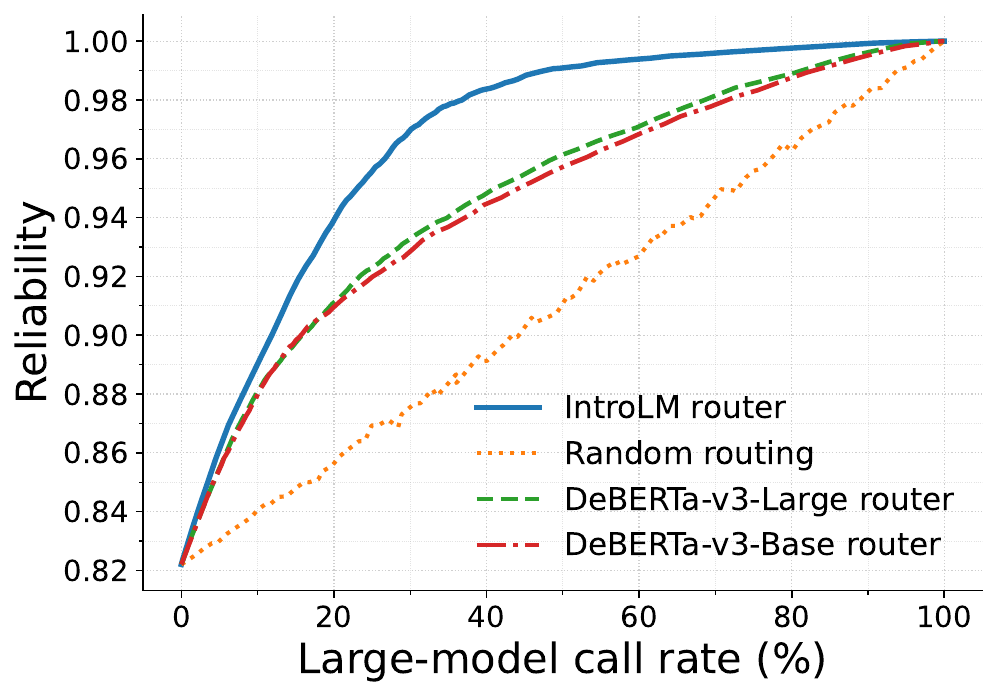}
        \caption{\textsc{Gen.\ QA:} Call rate}
        \label{fig:generalqa_rel_call}
    \end{subfigure}
    \hfill
    \begin{subfigure}[t]{0.24\textwidth}
        \centering
        \includegraphics[width=\linewidth]{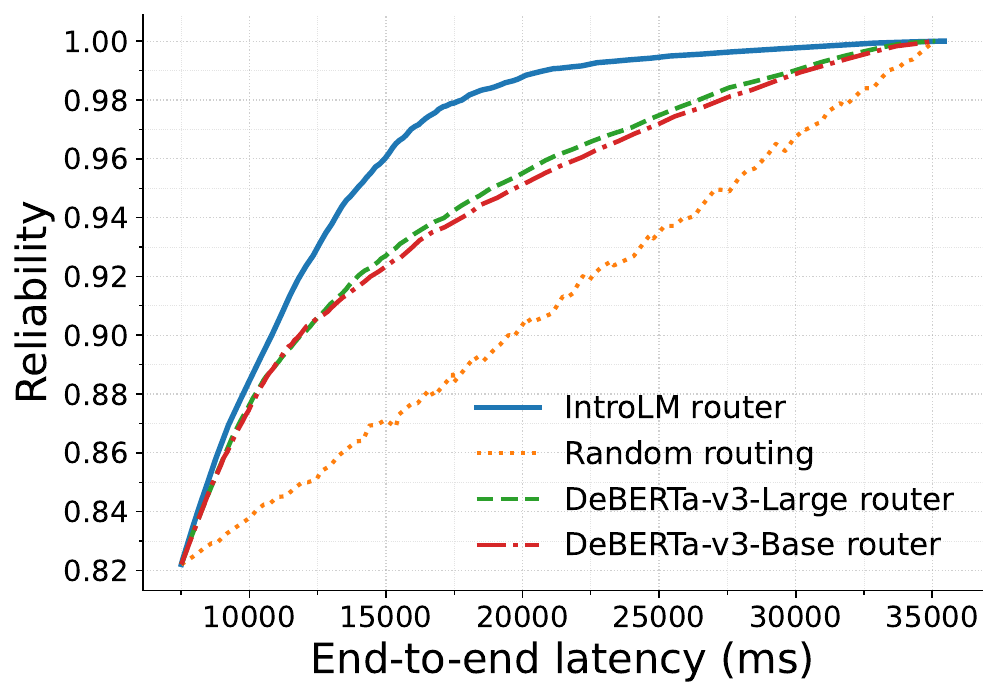}
        \caption{\textsc{Gen.\ QA:} Latency}
        \label{fig:generalqa_rel_lat}
    \end{subfigure}
    \hfill
    \begin{subfigure}[t]{0.24\textwidth}
        \centering
        \includegraphics[width=\linewidth]{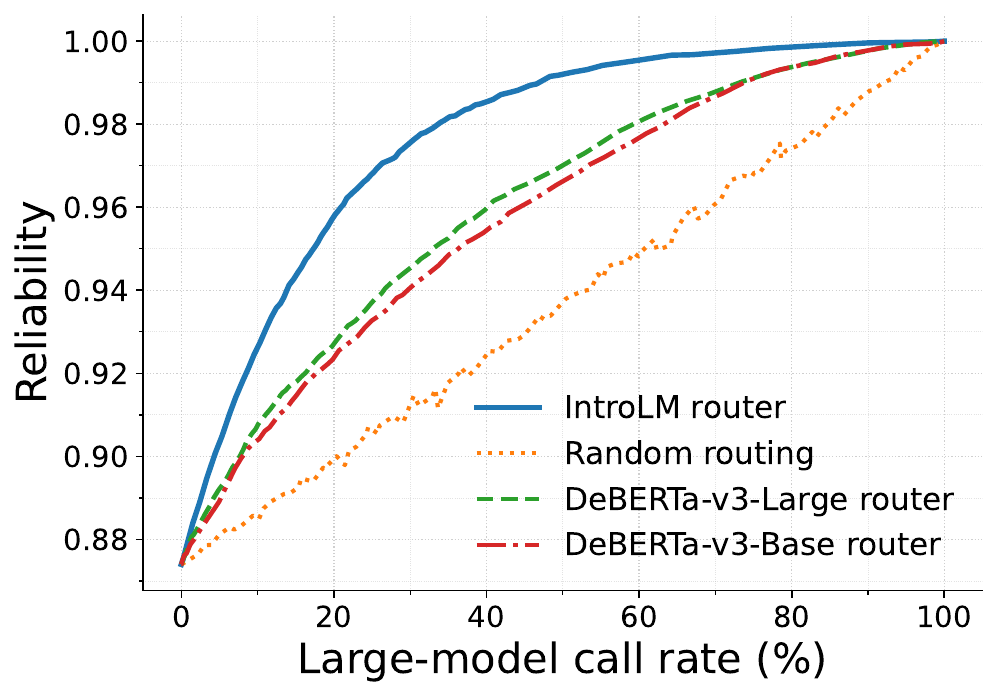}
        \caption{\textsc{HotpotQA:} Call rate}
        \label{fig:hotpotqa_rel_call}
    \end{subfigure}
    \hfill
    \begin{subfigure}[t]{0.24\textwidth}
        \centering
        \includegraphics[width=\linewidth]{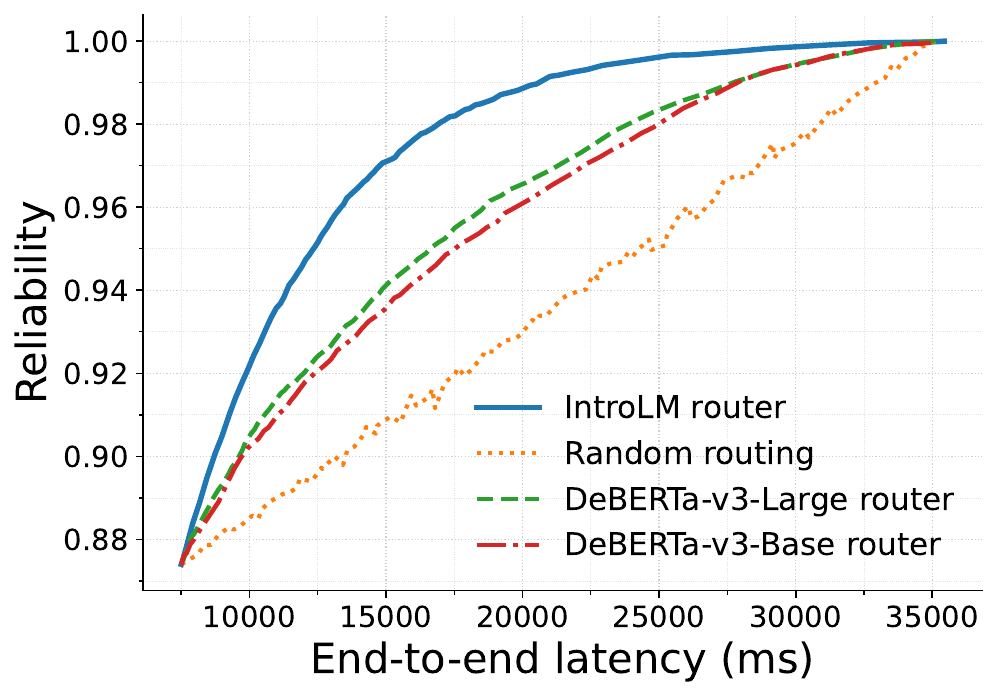}
        \caption{\textsc{HotpotQA:} Latency}
        \label{fig:hotpotqa_rel_lat}
    \end{subfigure}

    \caption{Routing trade-offs on \textsc{General QA} and \textsc{HotpotQA}. IntroLM (solid blue curve) consistently achieves lower
latency and fewer large-model invocations at matched reliability.}

    \label{fig:routing_tradeoffs_4panel}
\end{figure*}

\begin{table*}[t]
\centering
\small
\begin{tabular}{lcccccc}
\toprule
\textbf{Model} &
\multicolumn{2}{c}{\textsc{General QA}} &
\multicolumn{2}{c}{\textsc{HotpotQA}} &
\multicolumn{1}{c}{\textsc{Chat}} \\
\cmidrule(lr){2-3} \cmidrule(lr){4-5} \cmidrule(lr){6-6}
 & ROC--AUC & PR--AUC$_{\downarrow}$ & ROC--AUC & PR--AUC$_{\downarrow}$ & ROC--AUC \\
\midrule
DeBERTa-v3-Base (184M)  & 74.3 & 44.3 & 69.4 & 24.3 & 82.6 \\
DeBERTa-v3-Large (435M) & 75.8 & 45.5 & 71.8 & 26.8 & 86.3 \\
Matrix Factorization & 69.2 & 39.8 & 52.1 & 14.0 & 76.06 \\
IntroLM (Qwen3-8B)     & \textbf{89.1} & \textbf{63.4} & \textbf{86.3} & \textbf{46.7} & \textbf{90.1} \\
\bottomrule
\end{tabular}
\caption{Complexity prediction performance on structured question answering benchmarks
(\textsc{General QA: MMLU+MMLU-Pro+GSM8K}), long-context reasoning (\textsc{HotpotQA}), and daily-style
chat prompts (\textsc{LMSYS-Chat-1M}). We report ROC--AUC and PR--AUC for the complex
(negative) class where applicable; higher is better.}
\label{tab:complexity_eval}
\end{table*}

\paragraph{Models, Training, and Baselines}
For prompt complexity evaluation, we use \textsc{Qwen3-8B} \citep{yang2025qwen3} as the causal
language model backbone for IntroLM. As baselines, we consider two BERT-based classifiers,
\textsc{DeBERTa-v3-Base} and \textsc{DeBERTa-v3-Large} \citep{he2020deberta}, as well as a
matrix-factorization model adapted from prior routing work \citep{ongroutellm}. The
matrix-factorization baseline uses query embeddings together with a lightweight bilinear scoring
function to predict model success. For evaluating our IntroLM-based routing, we use the conventional
BERT-based routing system. A detailed explanation of models, training hyperparameters, and the
baseline routing system is available in Appendix~\ref{app:models}.

\subsection{Main Results}
We begin by evaluating IntroLM’s capability to predict prompt complexity and
compare it to BERT baselines. We then examine how these
complexity evaluators translate into end-to-end routing performance, analyzing
reliability, latency, and large-model call rate under different thresholds.

\subsubsection{Complexity Evaluation Performance}

We evaluate complexity prediction using \emph{ROC--AUC} and \emph{PR--AUC} as complementary, threshold-free metrics. By sweeping the decision threshold, ROC--AUC measures the trade-off between true and false positive rates, while PR--AUC reflects precision--recall trade-offs for rare complex prompts. Table~\ref{tab:complexity_eval} reports these results across datasets. IntroLM consistently outperforms all baselines, achieving substantial gains in both ROC--AUC and PR--AUC on the question-answering benchmarks. In addition to the BERT-based classifiers, we also compare against a matrix-factorization baseline adapted from prior routing work. This baseline performs poorly overall, particularly on \textsc{HotpotQA}, suggesting that lightweight embedding-based scoring methods are less effective for fine-grained complexity estimation in long-context reasoning settings. Figure~\ref{fig:roc_curves} plots the corresponding ROC curves and further shows that IntroLM achieves stronger separability than the competing approaches. While gains on chat-style datasets are smaller and the gap relative to \textsc{DeBERTa-v3-Large} is narrower, IntroLM remains the best-performing method overall. These results indicate that IntroLM is especially effective in reasoning-heavy and long-context settings, where simpler encoder-based or embedding-based baselines are less capable.

\subsubsection{Routing Performance}
We next evaluate how complexity evaluators translate into end-to-end routing
performance. Because routing decisions are most consequential on prompts with
substantial reasoning demands, we focus on the general question-answering (\textsc{General QA}) and
\textsc{HotpotQA} benchmarks, where inputs are inherently more complex and misrouting
incurs significant cost or accuracy penalties. Because routing systems are
deployed under different cost constraints, we consider two complementary
evaluation scenarios, each capturing a distinct practical setting.

\paragraph{Reliability vs.\ large-model call rate.}
In settings where the large model corresponds to a proprietary or high-cost
service, the primary objective is to minimize the fraction of prompts routed to
the large model while maintaining reliable output quality. To capture this
trade-off, we plot routing reliability as a function of the large-model call
rate. For each routing strategy, we vary the classification threshold
$\alpha$, and at each operating point measure (i) the resulting reliability and
(ii) the percentage of prompts escalated to the large model. This evaluation
directly reflects the effectiveness of a router in avoiding unnecessary
large-model invocations while preserving performance.

\paragraph{Reliability vs.\ latency.}
When both the small and large models are open-source and hosted locally,
end-to-end latency becomes the dominant cost factor. In this scenario, we
evaluate the trade-off between routing reliability and average latency. As
before, we sweep the threshold $\alpha$ and record the corresponding reliability
and end-to-end latency for each routing strategy. This analysis characterizes
how effectively a routing policy can reduce response time while maintaining a
desired level of output quality.
Together, these two views provide a comprehensive evaluation of routing
performance across cost-sensitive and latency-sensitive deployment scenarios.

\begin{table*}[t]
\centering
\small
\begin{tabular}{l l cc}
\toprule
\textbf{Trained Data} & \textbf{Model} & ROC--AUC & PR--AUC$_{\downarrow}$ \\
\midrule
\multirow{2}{*}{\textsc{General QA} (\textsc{Qwen3-8B})} 
 & \textsc{DeBERTa-v3-Large} & 75.8 & 45.5 \\
 & IntroLM (\textsc{Qwen3-8B}) & \textbf{89.1} & \textbf{63.4} \\
\midrule
\multirow{2}{*}{\textsc{General QA} (\textsc{Qwen3-1.7B})} 
 & \textsc{DeBERTa-v3-Large} & 75.68 & 64.71 \\
 & IntroLM (\textsc{Qwen3-1.7B}) & \textbf{84.24} & \textbf{72.72} \\
\midrule
\end{tabular}
\caption{Results on the effect of backbone model capacity on introspective
complexity evaluation. IntroLM consistently outperforms the BERT-based baseline
across both \textsc{Qwen3-8B} and \textsc{Qwen3-1.7B} settings, with stronger performance observed
for larger backbone models.}

\label{tab:complexity_eval_all}
\end{table*}

\begin{table*}[t]
\centering
\small
\begin{tabular}{l l cc}
\toprule
\textbf{Trained Data} & \textbf{Model} & ROC--AUC & PR--AUC$_{\downarrow}$ \\
\midrule
\multirow{2}{*}{\textsc{HotpotQA}} 
 & Backbone only (\textsc{Qwen3-8B}) & 81.0 & 35.8 \\
 & IntroLM (\textsc{Qwen3-8B}) & \textbf{86.3} & \textbf{46.7} \\
\midrule
\multirow{2}{*}{\textsc{General QA}} 
 & Backbone only (\textsc{Qwen3-8B}) & 87.1 & 59.3 \\
 & IntroLM (\textsc{Qwen3-8B}) & \textbf{89.1} & \textbf{63.4} \\
\midrule
\end{tabular}
\caption{Ablation study on the effect of \texttt{[CPX]} introspection tokens.
We compare IntroLM against a variant that applies the classifier to the final
hidden state of the last prompt token (``Backbone only'').}

\label{tab:cpx_effect}
\end{table*}

Figure \ref{fig:routing_tradeoffs_4panel} evaluates the end-to-end routing performance of IntroLM under both
cost-sensitive and latency-sensitive deployment scenarios. We measure inference
latency using vLLM \citep{kwon2023efficient} on two H100 GPUs, with \textsc{Qwen3-8B} as the small model and
\textsc{Qwen3-32B} as the large model, and compute end-to-end latency using the analytical
formulation in section~\ref{sec:metrics}. On the \textsc{General QA} benchmark, IntroLM reduces latency
by up to 34\% (15\% on average) while decreasing large-model calls by up to 50\%
(30\% on average) at matched reliability. On \textsc{HotpotQA}, IntroLM achieves up to
30\% latency reduction (18\% on average) and reduces large-model usage by up to
49\% (41\% on average), demonstrating consistent efficiency gains across both
datasets.

\subsection{Ablation Studies}
\label{app:ablations_main}
\subsubsection{Effect of Backbone Model Capacity}

We study how the capacity of the underlying language model affects introspective
complexity prediction. In addition to \textsc{Qwen3-8B}, we apply IntroLM to
\textsc{Qwen3-1.7B} using the same \textsc{General QA} dataset construction procedure,
where prompts are labeled based on whether the model’s generated output is
correct. Due to the reduced capability of \textsc{Qwen3-1.7B}, the resulting dataset is
more balanced in terms of success and failure labels. We also train a
\textsc{DeBERTa-v3-Large} classifier on the same \textsc{Qwen3-1.7B}-derived labels for comparison.

Table~\ref{tab:complexity_eval_all} shows that IntroLM with \textsc{Qwen3-1.7B} still
outperforms the BERT-based baseline, but achieves lower ROC--AUC and PR--AUC gains over the BERT baseline
than its \textsc{Qwen3-8B} counterpart. This trend indicates that increasing model
capacity improves the quality of introspective signals extracted during
prefilling, enabling more accurate prediction of a model’s own output quality.

\subsubsection{Effect of \texttt{[CPX]} Introspection Tokens }
\label{cpx_impact}
To isolate the impact of the proposed introspection tokens, we compare IntroLM against a variant that removes \texttt{[CPX]} tokens and applies the same classifier to the final hidden state of the last prompt token. Table \ref{tab:cpx_effect} shows the effect of IntroLM against a baseline backbone-only classifier. We observe particularly strong gains on \textsc{HotpotQA} when incorporating token-conditional LoRA, with the most pronounced improvements appearing in PR–AUC for the minority (failure) class. \textsc{HotpotQA} is inherently longer than our \textsc{General QA} benchmark, requiring multi-hop reasoning over long reference passages, with an average input length of approximately 1,500 tokens compared to roughly 300 tokens for the \textsc{General QA} dataset. Also, the class imbalance is higher in \textsc{HotpotQA}. In this setting, complexity arises not only from the question itself but also from processing extended contextual information during prefilling. The larger gains on \textsc{HotpotQA} indicate that token-conditional LoRA enables IntroLM to better adapt its introspective representations under the scenarios where distinguishing between complex and simpler prompts is more challenging, aligned with the current state of LLMs.

We also conducted a series of complementary ablation studies analyzing token-conditional LoRA
configurations, layer-wise introspection, and the effect of prefix truncation, available in Appendix~\ref{app:ablations}. Although we focus on binary routing in this work, IntroLM naturally extends to multi-model routing via shared \texttt{[CPX]} representations and model-specific classifier heads; preliminary results are provided in Appendix~\ref{app:multimodel_routing}.

\section{Conclusion}

We introduced IntroLM, a method that enables causal language models to predict their output quality during the prefilling phase without interfering with generation. By appending a \texttt{[CPX]} token and applying token-conditional LoRA, IntroLM learns complexity-aware representations while preserving the base model’s behavior and avoiding external classifiers. Across diverse question-answering benchmarks, IntroLM consistently outperforms strong BERT-based baselines in complexity prediction. When integrated into routing systems, it yields improved cost–performance trade-offs, reducing both latency and reliance on larger models at matched reliability.

\section{Acknowledgements}
We used OpenAI ChatGPT to assist with proofreading and improving the clarity and presentation of the manuscript.
\section{Limitations}

The primary objective of this work is to propose a general mechanism for prefilling-time prompt complexity evaluation that naturally accommodates arbitrary prompt lengths. While we evaluate IntroLM on a range of question-answering benchmarks with substantial reasoning demands, as well as a conventional chat-based benchmark, the space of LLM tasks has expanded considerably in recent years. Extending IntroLM to other task families—such as creative generation, multi-turn dialogue, code generation, or domain-specific applications—may require additional task-specific adaptation and evaluation, which we leave to future work.

Compared to lightweight BERT-based encoder classifiers, IntroLM incurs higher training-time computation, as it operates on the backbone language model and introduces token-conditional LoRA adapters rather than relying on a separate encoder. While IntroLM avoids full fine-tuning of the backbone and does not introduce additional overhead beyond what would be required to fine-tune a model of comparable size, it nonetheless remains more costly to train than conventional encoder-based classifiers. This trade-off reflects a deliberate design choice to support long-context inputs and prefilling-time introspection.

Our empirical study considers two backbone sizes and demonstrates that IntroLM achieves strong predictive performance in both settings. Although increased model capacity is generally expected to yield equal or improved introspective accuracy, a systematic evaluation of IntroLM on substantially larger backbones remains an important direction for future work, particularly in the context of emerging ultra-long-context models.

Finally, similar to much prior work on routing and complexity estimation, our approach relies on supervised signals derived from labeled datasets to train the complexity evaluator. While we construct these labels using established benchmarks, the framework could be extended to alternative supervision sources, such as preference-based training or other forms of weak or implicit feedback. Exploring such supervision paradigms is an interesting avenue for future research.

\bibliography{custom}

\appendix

\section{Related Work}
\label{app:related}
Output quality estimation has been widely used in routing studies, where the typical objective is to select the most efficient model to run the prompt and avoid using high-cost models. In this regard, studies like FrugalGPT \citep{chenfrugalgpt} and AutoMix \citep{automix} allow running the prompt on multiple models. FrugalGPT runs the prompt sequentially on a cascade of models, stopping when the output meets the desired quality. In contrast, AutoMix uses the same LLM that has generated the first response to verify it. However, doing the prefilling and decoding phases on multiple models to generate an output can add substantial cost and latency overheads to the system.  

Another line of work performs routing decisions before running the prompt on any model. HybridLLM \citep{hybridllm} and BEST-Route \citep{best-route} rely on BERT-style encoder classifiers to select between models, while RouteLLM \citep{ongroutellm} explores multiple routing strategies, including similarity-based ranking, matrix factorization, BERT classifiers, and causal LLM-based classifiers. ZOOTER \citep{zooter} similarly performs pre-generation routing by distilling reward-model signals into a lightweight encoder-based router. While these approaches avoid executing multiple models at inference time, they fundamentally depend on external evaluators or BERT-style classifiers with fixed and limited context windows. As a result, they are ill-suited for modern long-context settings—such as retrieval-augmented generation—where prompt complexity arises from processing substantial prefilling inputs, motivating approaches that leverage causal language models directly for complexity evaluation.

Among routing studies, \citep{confidence} shares some similarities with our work. They introduce two new tokens, <CN> (confident) and <UN> (unconfident). Then, they train the backbone to generate one of these tokens at the end of the decoding phase, to infer if the model is confident in its output. The key distinction between their work and ours is that they generate these tokens at the end of the generation (decoding phase). Hence, confidence evaluation is done after the high-latency decoding phase. However, IntroLM is applied to the prefilling phase to evaluate the output quality before any generation. This facilitates using causal language models for classification. Moreover, despite their work, we preserve the orthogonal weights for the generative phase.

\section{Dataset Construction and Labeling}
\label{app:datasets}
We provide additional details on dataset construction, preprocessing,
and labeling procedures used in our experiments.

\subsection{Datasets}

\paragraph{\textsc{General Question Answering Datasets.}}

We combine \textsc{MMLU} \citep{mmlu}, \textsc{MMLU-Pro} \citep{mmlu-pro}, and \textsc{GSM8K} \citep{gsm8k} into a unified
question-answering corpus covering general knowledge, adversarial questions, and
multi-step numerical reasoning. Each example consists of a question and a
ground-truth answer. We formed a diverse
question–answering corpus of 136,515 total questions here.

Exact-match evaluation can be brittle for generative models, as correct answers
may differ from ground-truth annotations due to minor formatting variations or
paraphrasing. To mitigate this issue, we use a lightweight LLM-as-judge to assess
semantic correctness. Specifically, we employ \textsc{LLaMA-3.1-8B-Instruct} to
determine whether the model’s response refers to the same entity or value as the
ground truth, allowing for minor surface-form differences. For this dataset, 21\% of the prompts were labeled complex (negative).

\begin{figure}[t]
\centering
\begin{tcolorbox}[
  colback=white,
  colframe=black,
  boxrule=0.5pt,
  arc=2pt,
  left=4pt,
  right=4pt,
  top=4pt,
  bottom=4pt
]
\small
\textbf{System:} You are evaluating whether a model’s answer to a question is
correct. Respond only in JSON format with fields
\texttt{is\_correct} (true/false) and \texttt{reason} (short explanation).

\vspace{4pt}
\textbf{Question:} \\
\textit{\{question\}}

\vspace{4pt}
\textbf{Ground-Truth Answer:} \\
\textit{\{ground\_truth\}}

\vspace{4pt}
\textbf{Model Answer:} \\
\textit{\{prediction\}}

\vspace{4pt}
\textbf{Instruction:} Treat the model’s answer as correct if it refers to the
same entity or value as the ground truth, allowing for minor formatting or
phrasing differences. Otherwise, mark it as incorrect.
\end{tcolorbox}
\caption{Judge prompt used for semantic correctness evaluation on question
answering datasets.}
\label{fig:qa_judge_prompt}
\end{figure}

\paragraph{\textsc{HotpotQA.}}

\textsc{HotpotQA} \cite{hotpotqa}, consisting of 97,074 samples, is used to evaluate complexity prediction under long-context,
reference-heavy inputs consisting of multiple evidence passages and a multi-hop
question. This setting places greater emphasis on input length, context
integration, and multi-step reasoning, and closely reflects the post-retrieval
reasoning phase of long-context retrieval-augmented LLM systems.

We follow standard preprocessing and apply the same model-based correctness
evaluation as in general question answering, using the prompt in
Figure~\ref{fig:qa_judge_prompt} to derive success labels. For \textsc{HotpotQA}, 14\% of the prompts were labeled complex (negative).

\paragraph{\textsc{LMSYS-Chat-1M.}}

To evaluate complexity prediction under realistic, open-ended conversational
inputs, we use \textsc{LMSYS-Chat-1M}, a large-scale corpus of real-world chat prompts.
Since the dataset consists of multi-turn conversations, we extract individual
user turns and treat each as an independent prompt, discarding dialogue history
to match our prefilling-time evaluation setting. We randomly sample 100K English
user prompts and remove trivial or context-dependent inputs.

\textsc{LMSYS-Chat-1M} provides no gold correctness labels and responses are inherently
less structured. We therefore employ a stronger judge,
\textsc{Qwen2.5-32B-Instruct}, to evaluate response quality. The judge assigns a
score from 0 to 10 based on relevance, accuracy, completeness, clarity, and
helpfulness. We convert scores into binary success labels by marking responses
with a score below 8 as unsuccessful. This is a design choice and can be changed based on the desired performance. Using this setting, approximately half of the prompts were labeled complex (negative).

\begin{figure}[t]
\centering
\begin{tcolorbox}[
  colback=white,
  colframe=black,
  boxrule=0.5pt,
  arc=2pt,
  left=4pt,
  right=4pt,
  top=4pt,
  bottom=4pt
]
\small
\textbf{System:} /no\_think You are an expert evaluator of conversational AI
responses.

\vspace{4pt}
\textbf{User Prompt:} \\
\textit{\{prompt\}}

\vspace{4pt}
\textbf{Assistant Response:} \\
\textit{\{response\}}

\vspace{4pt}
\textbf{Instruction:} Evaluate the response based on relevance, accuracy,
completeness, clarity, and helpfulness. Return a JSON object containing a
numerical \texttt{score} from 0 (very poor) to 10 (excellent), along with a brief
justification.
\end{tcolorbox}
\caption{Judge prompt used for model-based evaluation of chat-style responses in
\textsc{LMSYS-Chat-1M}.}
\label{fig:chat_judge_prompt}
\end{figure}

\subsection{Data Splits}

All datasets use a consistent 80/10/10 train/validation/test split. Splits are
constructed at the prompt level and fixed across all experiments.

\section{Models, Training, and Baselines}
\label{app:models}
\paragraph{Complexity Evaluation}For IntroLM, we use \textsc{Qwen3-8B} \citep{yang2025qwen3} as the causal
language model and append the \texttt{[CPX]} token for complexity
evaluation. Training uses a context window of 2048 tokens, a batch size of~64, cosine learning-rate scheduling
with a 10\% warm-up ratio, and class-weighted binary cross-entropy. Token-
conditional LoRA (rank~32, $\alpha=64$) is applied  to
\texttt{q\_proj}, \texttt{o\_proj}, \texttt{gate\_proj}, \texttt{up\_proj},
and \texttt{down\_proj}. We use a maximum gradient norm of~0.3, weight decay of
0.002, and learning rates in the range of $4$--$8\times 10^{-5}$ for the
classifier, aggregator, embeddings, and LoRA parameters.

For the BERT-based baselines, we use \textsc{DeBERTa-v3-Base} and
\textsc{DeBERTa-v3-Large} \citep{he2020deberta}, representing a strong encoder
architecture at two different capacity levels. Both models operate on the
\texttt{[CLS]} representation with a linear classifier and are trained using a
batch size of~16, class-weighted binary cross-entropy, AdamW with learning rate
$3\times 10^{-5}$, weight decay~0.01, and dropout~0.1. All baselines are trained
on the same datasets and label definitions as IntroLM to ensure a fair
comparison.

We also include a lightweight embedding-based baseline adapted from the matrix-factorization router of \citet{ongroutellm}. Unlike encoder-based or generative classifiers, this method operates directly on frozen prompt embeddings and therefore has very low modeling overhead. Each prompt is encoded using the \textsc{all-MiniLM-L6-v2} sentence-transformer model \citep{reimers2019sentencebert, allminilm_l6_v2} to obtain a 384-dimensional representation, which is then passed to a lightweight bilinear scorer that predicts the probability that the target model will successfully answer the prompt. In our binary success-prediction setting, this corresponds to a simplified single-route variant of the original matrix-factorization formulation, trained with binary cross-entropy on the same instance-level labels as IntroLM and the BERT baselines. We optimize the model using AdamW with a learning rate of $3 \times 10^{-4}$ and a batch size of 64.
\paragraph{Prompt Routing}
We consider conventional BERT-based routing systems as the baseline for evaluating our IntroLM-based routing approach. We consider pre-routing strategy based on an
external classifier. A BERT-based model
$g_{\text{BERT}} : \mathcal{X} \rightarrow [0,1]$ operates on the
\texttt{[CLS]} representation of the prompt and predicts the probability that
$M_s$ can successfully handle the input. We follow the same thresholding
rule as IntroLM-based routing (see Section~\ref{threshold}), but decisions are made \emph{prior} to any LLM computation. This
baseline represents a common approach in prior work and serves as a point of
comparison for evaluating the benefits of prefill-aware routing with IntroLM.

\section{Ablation Studies}
\label{app:ablations}

\begin{table}[t]
\centering
\small
\begin{tabular}{lcc}
\toprule
\textbf{LoRA Target Modules} & ROC--AUC & PR--AUC$_{\downarrow}$ \\
\midrule
FFN only        & 89.1 & 63.0 \\
Attention only & 88.5 & 61.0 \\
No LoRA & 85.7 & 56.4 \\
Attention + FFN (full)                   & \textbf{89.1} & \textbf{63.1} \\
\bottomrule
\end{tabular}
\caption{Ablation of token-conditional LoRA target modules for complexity
evaluation on the \textsc{General QA} benchmark with \textsc{Qwen3-8B}. We report ROC--AUC and
PR--AUC for the complex (negative) class (higher is better).}

\label{tab:lora_modules_ablation}
\end{table}

\subsection{Effect of Token-Conditional LoRA}
\label{LoRA_effects}
We examine four configurations of token-conditional adaptation, summarized in
Table~\ref{tab:lora_modules_ablation}. As shown, applying token-conditional LoRA
only to the feed-forward (FF) projections yields performance comparable to
adapting both the attention and FF projections. Restricting LoRA to the
attention projections (\texttt{q\_proj}, \texttt{o\_proj}) results in a modest
drop in accuracy. In contrast, when token-conditional LoRA is not used and only
the \texttt{[CPX]} embedding and classifier head are trainable, performance
degrades substantially, with a pronounced reduction in PR--AUC for the complex
(negative) class. This indicates that, without token-conditional adaptation, the
model struggles to reliably identify escalation-worthy prompts, highlighting
the importance of LoRA-based adaptation for effective complexity evaluation.

\begin{table}[t]
\centering
\small
\begin{tabular}{lcc}
\toprule
\textbf{Model / Layers Used} & ROC--AUC & PR--AUC$_{\downarrow}$ \\
\midrule
\textsc{Qwen3-1.7B} (first 18 / 28) & 83.3 & 71.5 \\
\textsc{Qwen3-1.7B} (first 22 / 28) & \textbf{84.2} & 72.5 \\
\textsc{Qwen3-1.7B} (full model) & \textbf{84.2} & \textbf{72.7} \\
\midrule
\textsc{Qwen3-8B} (first 18 / 36) & 81.3 & 51.0 \\
\textsc{Qwen3-8B} (first 24 / 36) & 87.9 & 60.1 \\
\textsc{Qwen3-8B} (full model) & \textbf{89.1} & \textbf{63.0} \\
\bottomrule
\end{tabular}
\caption{Ablation of layer-wise introspection for IntroLM. The classifier head
and token-conditional LoRA are applied to intermediate layers rather than the
full backbone. We report ROC--AUC and PR--AUC for the complex (negative) class on
the \textsc{General QA} benchmark.}
\label{tab:layerwise_ablation}
\end{table}

\subsection{Layer-wise introspection.}
\label{app:layerwise_intro}
We study whether IntroLM can produce reliable complexity estimates when the
classifier head and token-conditional LoRA are applied to intermediate layers
rather than the full backbone. This enables earlier complexity signals during
prefilling and allows training on only a prefix of the model, substantially
reducing memory and compute—an appealing property for very large language
models. We evaluate this setting on \textsc{Qwen3-1.7B} and \textsc{Qwen3-8B} by attaching the
classifier and LoRA to progressively deeper layer prefixes. As shown in
Table~\ref{tab:layerwise_ablation}, performance improves monotonically with
depth, approaching full-model accuracy while using only a subset of layers.
These results indicate that meaningful introspective signals emerge well before
the final layer, and that deeper backbones benefit more strongly from later-layer
introspection.

\subsection{Effect of Prefix Truncation}
\label{app:prefix_truncation}

A practical question for prefilling-time introspection is whether reliable success-prediction signals can be obtained before processing the full input sequence. This is particularly relevant in long-context settings, where one may wish to make routing decisions earlier by truncating the prompt prefix rather than waiting for the full prefilling pass to complete.

\begin{table}[t]
\centering
\small
\begin{tabular}{lcc}
\toprule
Approach & ROC--AUC & PR--AUC \\
\midrule
DeBERTa-v3-Large & 71.8 & 26.8 \\
IntroLM (full context) & 86.3 & 46.7 \\
IntroLM (first 512 tokens) & 77.8 & 33.1 \\
\bottomrule
\end{tabular}
\caption{Effect of prefix truncation on complexity prediction for \textsc{HOTPOTQA}. Truncating the input to the first 512 tokens noticeably reduces performance relative to full-context IntroLM, indicating that access to the full prompt is important for reliable success prediction in long-context reasoning tasks.}
\label{tab:prefix_truncation}
\end{table}

To study this trade-off, we evaluate IntroLM on \textsc{HOTPOTQA} under prefix truncation. In addition to the standard full-context setting, we train a variant of IntroLM on only the first 512 input tokens using the same \textsc{Qwen3-8B} backbone. We compare both variants against the \textsc{DeBERTa-v3-Large} baseline. Table~\ref{tab:prefix_truncation} summarizes the results.

The results show that truncating the prompt to the first 512 tokens leads to a clear degradation in performance relative to the full-context version of IntroLM. While the truncated variant still outperforms the BERT-based baseline, it falls substantially short of full-context introspection, especially on PR--AUC. This suggests that, for long-context reasoning tasks such as \textsc{HOTPOTQA}, reliable success prediction depends on information distributed across the full prompt rather than only its initial prefix.

These findings complement the layer-wise introspection results in Appendix~\ref{app:layerwise_intro}, which show that useful prediction signals can be extracted from intermediate layers without waiting for the final layer. Taken together, the results indicate that earlier prediction may be achieved more effectively through intermediate-layer introspection over the full sequence than through aggressive prefix truncation. A broader investigation of chunked or progressive routing decisions over long contexts is an important direction for future work.

\section{Preliminary Results on Multi-Model Routing}
\label{app:multimodel_routing}

While our main experiments focus on the binary routing setting, IntroLM is not inherently restricted to selecting between only two models. The method naturally extends to multi-model routing by attaching multiple lightweight classifier heads to the shared \texttt{[CPX]} representation, where each head predicts the success probability of a different candidate model. Routing can then be performed by selecting among candidate models based on these predicted probabilities together with the desired cost, latency, or reliability objective.

\begin{table}[t]
\centering
\small
\begin{tabular}{lcc}
\toprule
Backbone Used for Prediction & ROC--AUC & PR--AUC \\
\midrule
Qwen3-8B & 83.76 & 72.36 \\
Qwen3-1.7B & 84.24 & 72.72 \\
DeBERTa-v3-Large & 75.68 & 64.71 \\
\bottomrule
\end{tabular}
\caption{Preliminary results for extending IntroLM beyond binary routing. We train predictors for the success labels of \textsc{Qwen3-1.7B}. An IntroLM model built on \textsc{Qwen3-8B} predicts \textsc{Qwen3-1.7B} performance nearly as well as the \textsc{Qwen3-1.7B} backbone itself, and substantially better than a BERT-based classifier.}
\label{tab:multimodel_extension}
\end{table}

This extension is analogous to prior multi-model routing approaches based on a shared encoder with model-specific prediction heads, but here the shared representation is produced by the causal language model itself during prefilling. Since the \texttt{[CPX]} token aggregates prompt-conditioned introspective signals from the full input context, it provides a natural representation for estimating the capability of multiple candidate models without requiring separate encoder-based evaluators.

As a preliminary validation of this setting, we train an IntroLM model built on a \textsc{Qwen3-8B} backbone to predict whether a smaller \textsc{Qwen3-1.7B} model would answer a prompt correctly. We compare this against two baselines: (i) an IntroLM model built on the \textsc{Qwen3-1.7B} backbone and trained to predict its own success labels, and (ii) a \textsc{DeBERTa-v3-Large} classifier trained on the same target labels. Table~\ref{tab:multimodel_extension} shows that \textsc{Qwen3-8B} predicts the success of \textsc{Qwen3-1.7B} nearly as well as the \textsc{Qwen3-1.7B} backbone itself, and substantially better than the BERT-based baseline.

These results suggest that IntroLM's prefilling-time representations are sufficiently informative not only for self-evaluation, but also for estimating the capability of other candidate models. This supports the feasibility of extending IntroLM to multi-model routing using a shared backbone representation with model-specific classifier heads. A full investigation of multi-model routing is left to future work.

\end{document}